\documentclass[letterpaper, 10 pt, conference]{ieeeconf}  

\IEEEoverridecommandlockouts                              

\usepackage{cite}
\usepackage{amsmath,amssymb,amsfonts}
\usepackage{adjustbox}  
\usepackage{subcaption}
\usepackage{multirow}
\usepackage{graphicx}
\usepackage{hyperref}

\newcommand{\orcidicon}{\includegraphics[width=0.32cm]{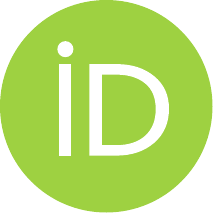}}
\expandafter\xdef\csname orcidCT\endcsname{\noexpand\href{https://orcid.org/\csname orcidauthorCT\endcsname}{\noexpand\orcidicon}}
\expandafter\xdef\csname orcidYG\endcsname{\noexpand\href{https://orcid.org/\csname orcidauthorYG\endcsname}{\noexpand\orcidicon}}
\expandafter\xdef\csname 
orcidCR\endcsname{\noexpand\href{https://orcid.org/\csname orcidauthorCR\endcsname}{\noexpand\orcidicon}}
\expandafter\xdef\csname orcidJG\endcsname{\noexpand\href{https://orcid.org/\csname orcidauthorJG\endcsname}{\noexpand\orcidicon}}

\title{\LARGE \bf
Autonomous Hiking Trail Navigation via Semantic Segmentation and Geometric Analysis}

\author{Camndon Reed$^{1}$\orcidCR{}, Christopher Tatsch$^{2}$\orcidCT{}, Jason N. Gross$^{2}$\orcidJG{} and Yu Gu$^{2}$\orcidYG{}
\thanks{*This research was supported in part by NSF Award number 2348288 “REU Site: Undergraduate Robotics Research for Rural Appalachia”}
\thanks{$^{1}$The authors are with the Department of Engineering, James Madison University, Harrisonburg, VA 22801, USA }
\thanks{$^{2}$The authors are with the Department of Mechanical, Materials and Aerospace Engineering, West Virginia University, Morgantown, WV 26505, USA 
        {\tt\small christophertatsch@ieee.org; yugu@mail.wvu.edu}}%
}

\begin{document}

\maketitle
\thispagestyle{empty}
\pagestyle{empty}


\begin{abstract}
Natural environments pose significant challenges for autonomous robot navigation, particularly due to their unstructured and ever-changing nature. Hiking trails, with their dynamic conditions influenced by weather, vegetation, and human traffic, represent one such challenge. This work introduces a novel approach to autonomous hiking trail navigation that balances trail adherence with the flexibility to adapt to off-trail routes when necessary. The solution is a Traversability Analysis module that integrates semantic data from camera images with geometric information from LiDAR to create a comprehensive understanding of the surrounding terrain. A planner uses this traversability map to navigate safely, adhering to trails while allowing off-trail movement when necessary to avoid on-trail hazards or for safe off-trail shortcuts. The method is evaluated through simulation to determine the balance between semantic and geometric information in traversability estimation. These simulations tested various weights to assess their impact on navigation performance across different trail scenarios. Weights were then validated through field tests at the West Virginia University Core Arboretum, demonstrating the method's effectiveness in a real-world environment.

\end{abstract}

\section{Introduction}
\label{introduction}

Hiking trails serve as gateways to nature, offering outdoor enthusiasts across the globe access to diverse landscapes and ecosystems. In the United States alone there are over 193,500 miles of trails on federal lands, with state and local trails adding significantly to this figure \cite{AHS2015}. The popularity of hiking continues to grow, with approximately 58.7 million participants in 2021 \cite{OIA2022}.  While these trails provide invaluable recreational opportunities, they also present unique problems for management, environmental conservation, and emergency response. Autonomous vehicles capable of navigating hiking trails could address these challenges effectively, providing valuable tools including trail monitoring, guided scenic tours, and trail maintenance. They also can support conservation efforts with wildlife tracking and ecosystem assessment, and improve hiker safety with search and rescue operations.



The unstructured and challenging nature of hiking trails poses significant difficulties for current autonomous navigation systems. Unlike structured urban environments with well-defined roadways and clear landmarks, hiking trails present a myriad of complexities. These paths are characterized by irregular terrain, inconsistent vegetation coverage, and unpredictable obstacles. Vegetation along trails varies greatly, ranging from thick wooded areas to sparse clearings. This variance complicates feature detection and can impact sensor performance over time. Moreover, hiking trails are dynamic, and constantly changing due to weather conditions, seasonal variations, and natural processes such as erosion and plant growth. 

\begin{figure}
\centering
\includegraphics[width=0.8\linewidth]{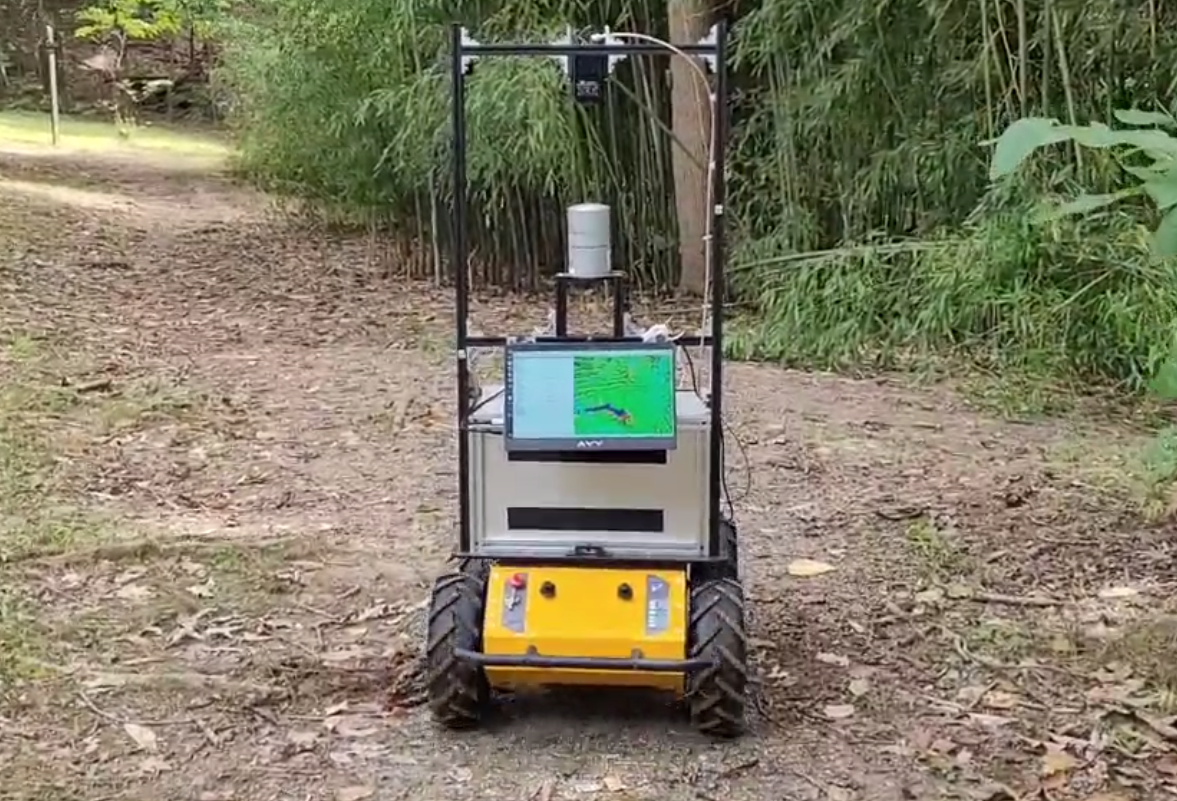}
\caption{Robot driving on the arboretum trail.} 
\label{fig:robot_on_trail}
\end{figure}   


The primary goal of this work is to improve autonomous hiking trail navigation for wheeled unmanned ground vehicles by balancing three key factors: adherence to the trail, off-trail shortcut identification, and avoidance of hazards. To achieve this balance, we developed a terrain traversability analysis module that integrates both semantic and geometric information, providing a more comprehensive understanding of the trail environment. Figure \ref{fig:robot_on_trail} shows the testbed robot, BrambleBee \cite{ohi2018design} driving on a hiking trail. Additionally, we developed a waypoint selection module that selects intermediate goals to navigate trails under normal scenarios while re-routing off-trail to avoid unexpected obstacles and take shortcuts when it is safe to do so allowing the system to understand better and respond to the complexities of trail environments. 
Our contributions are as follows:
\begin{enumerate}
    \item A traversability analysis module for combining the LiDAR\- and stereo camera maps containing geometric and semantic information in real-time.
    
    \item A waypoint selection module that balances on-trail navigation with off-trail detours. Our module prioritizes adhering to established trails while identifying safe shortcuts and avoiding hazards on and off the trail.
    \item An open-source dataset of semantically labeled trail images and a simulation environment to support further research and development in autonomous hiking trail navigation \footnote{\href{https://ieee-dataport.org/documents/hiking-trail-semantic-segmentation-image-dataset}{Hiking Trail Semantic Segmentation Images and Simulation dataset}}
    \item We demonstrate our method using a ground robot on actual hiking trails.
\end{enumerate}

The remainder of this paper is structured as follows: Section \ref{related_work} provides a review of related work in terrain classification, traversability analysis, and waypoint selection. Section \ref{methodology} describes our methodology, including the development of the traversability analysis module, global planner, and the simulation environment. Section \ref{experiments_results} presents the results of our field tests and discusses the effectiveness of our approach in real-world scenarios. Finally, Section \ref{conclusion} concludes the paper and suggests directions for future research in autonomous hiking trail navigation.

\section{Related Work}
\label{related_work}
Understanding the environment around the robot is at the core of safe and efficient navigation of hiking trails. Terrain traversability analysis, as defined in \cite{s21010073}, is the process of estimating the difficulty of driving through a terrain. This involves assessing the safety of potential paths by evaluating features such as roughness, slope, and terrain type. There are several approaches to this problem, including geometric-based methods, semantic-based methods, and integrated methods.  

Geometric-based methods rely on 3D point cloud data typically acquired through LiDAR sensors to analyze terrain shape and structure. These techniques employ various metrics such as slope, roughness, and surface normal estimation to determine traversability \cite{PAPADAKIS20131373}. While effective for detecting larger obstacles such as tree trunks or boulders, these methods can struggle with smaller trail hazards that are critical for safe navigation \cite{Borges2022ASO}. In \cite{9721819}, researchers proposed a terrain traversability estimation technique using LiDAR. This method faces challenges with irregular features such as tall grass or dense bushes, which can be misclassified as obstacles. Attempting to address the limitation caused by irregular ground features, \cite{article} introduces a LiDAR-only traversability estimation module designed for heavily vegetated environments. This approach employs a 3D voxel binary representation of the environment. However, the method’s effectiveness in real-world scenarios is unverified. Additionally, its binary classification system either limits the robot's navigation options if classification is too conservative or could put the robot in a dangerous position if careless.


Semantic-based approaches to traversability estimation leverage computer vision and machine learning techniques to understand and classify the content of a scene \cite{Borges2022ASO}. Semantic segmentation involves categorizing each pixel in an image into classes such as ‘path’, ‘tree’, or ‘rock’ \cite{Guo2017ARO}.  These methods typically employ Convolutional Neural Networks (CNNs) such as Mask R-CNN \cite{he2017mask} and TensorMask \cite{chen2019tensormask}. Presented in \cite{dabbiru2021} is a binary traversability mapping approach that leverages 2D camera data to identify traversable and non-traversable terrain in off-road environments. Off-Road Semantic Segmentation (OFFSEG), presented in \cite{Viswanath2021OFFSEGAS}, builds on these concepts by offering a more detailed framework for off-road semantic segmentation. Unlike binary systems, this approach enhances scene understanding by classifying terrain into sky, traversable, non-traversable, and obstacles, while also identifying specific sub-classes within identified traversable regions, such as grass, puddle, dirt, and gravel. However, OFFSEG and similar semantic-based methods are inherently limited due to their reliance on visual data, which limits their use in varying weather conditions and low-visibility scenarios. 



Recent research has focused on integrating semantic and geometric information to improve traversability estimation. In \cite{maturanadaniel}, the authors propose a $2.5$D grid map encoding both terrain height and navigation-relevant semantic classes such as trail and grass. In \cite{Leung}, they introduce a comparable representation, assigning each grid within the map a traversability cost. Although both of these methods show improvements over previous LiDAR-only systems in gauging safely traversable regions, they face limitations when encountering complex structures such as overhanging vegetation. This is due to the $2.5$D grid-based nature of their environment representations, which only store a single weight value for each $(x, y)$ coordinate. Such simplification can lead to incomplete modeling of multi-level features or vertical surfaces, potentially overlooking critical hazards within the environment. 



Past systems for waypoint selection have fused global information with the localization data such as vehicle kinematic and perception uncertainties of an agent at any given position \cite{vilca2016optimal}. This approach allows an agent to react in a predictive manner to situational circumstances without the need to perform computationally costly searches repeatedly \cite{vilca2016optimal}. Systems have also harnessed a sequence of several intermediate waypoints being placed relatively close together to guide an agent through tight spaces \cite{moreno2019automatic}. However, these implementations usually face limitations in terms of flexibility and adaptability to unstructured environments such as hiking trail networks. The role of semantic analysis in waypoint placement has been utilized to differentiate an environment between a roadway and off-road information based on GPS data \cite{zoccoler2015automatic}. This online approach faces limitations in a forest environment where service connection may be unsatisfactory. 

An effective strategy for planning a path in complex environments is a hierarchical approach such as in the TARE exploration algorithm \cite{cao2022autonomous}. In this method, a coarse initial path is generated using a Traveling Salesman Problem (TSP)-based solution derived from a global map, and then this path is locally optimized incorporating additional information. However, this method does not account for semantic information. The \cite{thoresen2021path} planning algorithm is designed specifically for using a traversability analysis map as input, however faces a similar limitation: incorporating a more detailed traversability map increases computational demands and requires a reduced planning horizon.


\section{Methodology}
\label{methodology}

\begin{figure}
\centering
\includegraphics[width=0.95\linewidth]{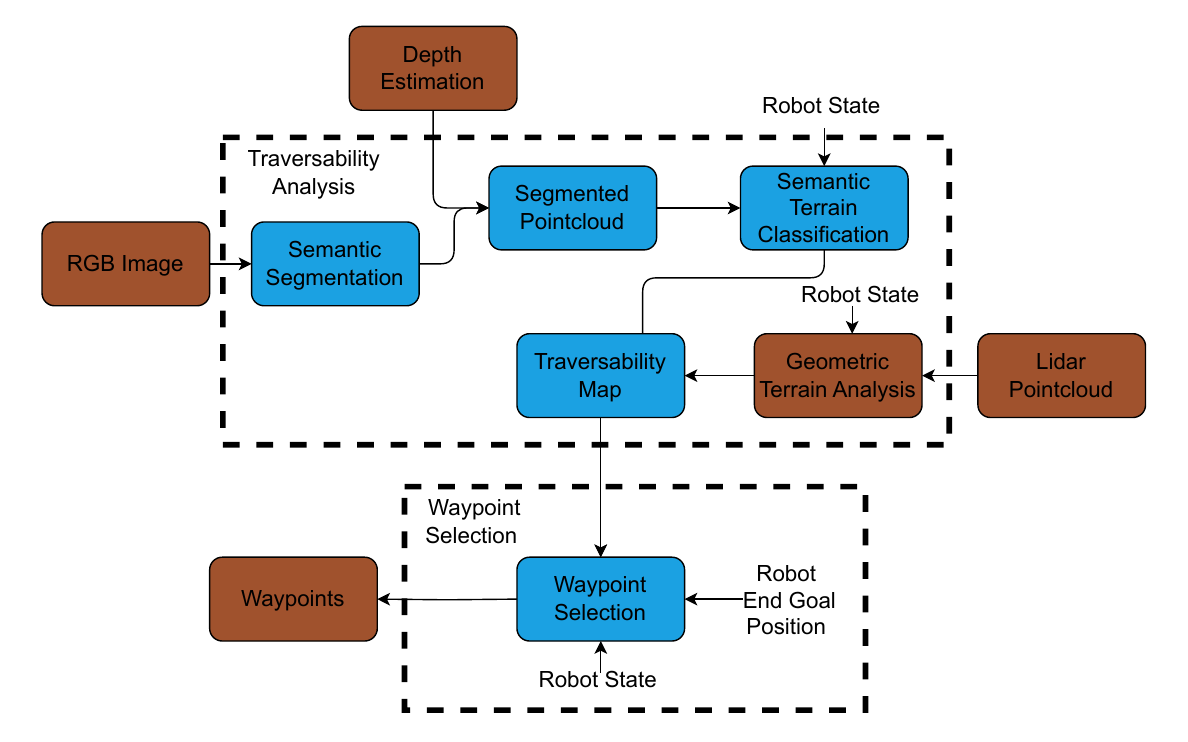}
\caption{System overview including the terrain analysis and the waypoint selection modules.} 
\label{fig:method_diagram}
\end{figure}

The primary goal of this work is to develop a system that can identify hiking trails and safely analyze the terrain traversability for a wheeled ground robot equipped with a depth camera and a LiDAR sensor. Once trail data is collected we also develop a framework for navigating this trail by planning intermediate waypoints that balance trail adherence, possible shortcuts, and obstacle avoidance. Our system consists of two main modules: traversability analysis and waypoint selection. The system architecture is shown in Figure \ref{fig:method_diagram}. The blue blocks are methods that are part of the developed modules, and the brown blocks are third-party modules required by the system to operate. 



The traversability analysis module integrates semantic and geometric information to generate a traversability map. The RGB images from a stereo camera undergo semantic segmentation using a YOLO \cite{RedmonDGF15} model trained on a custom hiking trail dataset. Simultaneously, the camera's depth estimation is used to generate point clouds. These point clouds are then labeled by projecting masks obtained from the segmentation results onto them, creating 3D representations with associated class labels. The semantic point clouds are fused with the geometric terrain analysis obtained from LiDAR data. Both information are fused based on a cost function. The registered point cloud result is a global traversability map representing the environment near the trail.

Given a global waypoint goal, the waypoint selection module generates intermediate waypoints that consider both geometric and semantic information from the traversability map; enabling the robot to follow hiking trails but also to be flexible in the case of encountering unexpected obstacles on the trail and enabling safe shortcuts. These local waypoints are sent to a local planner that considers a particular robot's kinematics and dynamics to follow the path.

\subsection{Traversability Analysis}

\begin{figure}[b]
\centering
\begin{subfigure}[b]{0.4\linewidth}
    \centering
    \includegraphics[width=\textwidth]{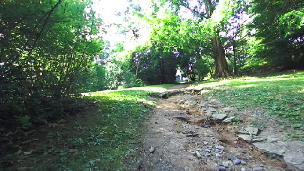}
    \caption{RGB image input.}\label{fig:fig_01_robot}
\end{subfigure}
\begin{subfigure}[b]{0.4\linewidth}
    \centering
    \includegraphics[width=\textwidth]{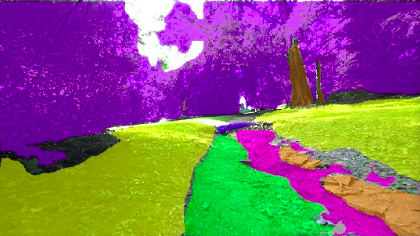}
    \caption{Segmentation results.}\label{fig:fig_01_segmentation}
\end{subfigure}

\begin{subfigure}[b]{0.5\linewidth}
    \centering
    \includegraphics[width=\textwidth]{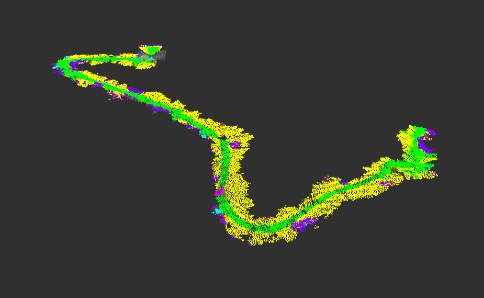}
    \caption{Semantic terrain classification point cloud.}\label{fig:fig_01_segmentation2}
\end{subfigure}
\caption{Semantic-based terrain classification generation.} 
\label{fig:traversability_analysis}
\end{figure} 

At the core of the traversability analysis module is a semantic segmentation model trained on a dataset of 1250 images, manually labeled into eight different classes: grass, rock, trail, root, structure, tree trunk, vegetation, and rough trail. The data set was split into training ($70\%$), validation ($20\%$), and test ($10\%$) sets. Roboflow’s \cite{Dwyer2024} polygon tool, which leverages the SegmentAnything Model \cite{kirillov2023segment}, was utilized to streamline the labeling process. A YOLOv8 \cite{Jocher_Ultralytics_YOLO_2023} segmentation model was trained on this dataset for one hundred epochs using a batch size of $64$. The following data augmentation techniques were utilized for training: rotations up to thirteen degrees in both directions, a four-degree shear in all four directions, and random noise covering up to $0.18\%$ of pixels in any given image. The model achieved a $75.8\%$ mean average precision (mAP) across all classes, with a notable $96.7\%$ mAP in identifying established trails. 

The first step in building the semantic map is inference. This is performed using the trained semantic segmentation YOLO model on the RGB image, generating binary masks that represent each semantic class in the image. Depth estimation data is then used to create a semantic 3D pointcloud. This conversion is done using the pinhole camera method, where for each pixel $(u, v)$ with depth $Z$, the corresponding 3D point $(X,Y,Z)$ is computed as $X = (u - c_x) \cdot Z / f_x $ and  $ Y = (v - c_y) \cdot Z / f_y$. Where $c_x$ and $c_y$ are the principal point offsets, and $f_x$ and $f_y$ are the focal lengths in pixels. The generated pointcloud undergoes initial filtering based on both distance from the robot and height relative to the robot. Points outside of the range $0.5m$ to $6.0m$ away from the camera are filtered out. This range was chosen due to the robot ZED2i camera depth estimation parameters, where depth accuracy is acceptable within this range. The robot's current attitude is taken from a robot localization estimation module and is used to filter out all points above the height of the agent. This step is helpful in reducing noise caused by both false depth estimation and excessive vegetation.

The semantic information obtained from the YOLO model is projected onto the point cloud by overlaying binary masks obtained from YOLO inference onto corresponding point clouds. Each point is thus assigned a semantic class label based on its spatial correspondence with segmented regions. To optimize the point cloud for further processing, two additional filtering steps are applied: 
\begin{enumerate}
    \item \textit{Voxel grid filtering}: This technique downsamples the pointcloud by dividing the 3D space into a grid of voxels of size $s$, with $s=0.1m$ being used in our implementation. For each voxel, all points within it are represented by thier centroid. For a voxel containing $n$ points $\{p_1, p_2, ..., p_n\}$, the centroid $c$ is calculated as:
$c = \frac{1}{n} \sum_{i=1}^n p_i$. This process significantly reduces the number of points while maintaining the overall structure of the point cloud.
    \item \textit{Statistical outlier removal}: This method identifies and removes points that are statistically distant from their neighbors. For each point, the mean distance $\mu$ to its $k$ nearest neighbors is computed. Points are considered outliers if their mean distance falls outside the range $[\mu_\text{global} - \alpha\sigma, \mu_\text{global} + \alpha\sigma]$, where $\mu_\text{global}$ is the mean of all mean distances, $\sigma$ is the standard deviation of all mean distances, and $\alpha$ is a user-defined threshold. $\alpha=2.0$ and $k=10$ were used in our implementation.
\end{enumerate}

Figure \ref{fig:traversability_analysis} illustrates this process and showcases classification results on the global scale using color coding. In the images, trail area is shown in green, grass in yellow, rocks in orange, tree trunks in brown, human made structure in cyan, roots in blue, vegetation in purple, and areas classified as rough-trail in pink.

Concurrently, geometric terrain analysis is performed on an elevation map obtained from a LiDAR pointcloud. Our approach leverages the method presented in \cite{cao2022autonomous} from Carnegie Mellon University (CMU), which provides a reliable solution for analyzing terrain geometry considering the terrain slope and height. This analysis generates a geometric terrain map where values above 0.1 are considered not traversable. 

Both pointclouds are transformed to a common map frame before association. A dual k-d tree data structure \cite{skrodzki2019kd} is used 
to combine points between the semantically segmented pointcloud and the geometric terrain analysis pointcloud. K-d trees are constructed for both pointclouds, enabling efficient bidirectional nearest-neighbor searches. For each point $p_t$ in the geometric terrain analysis pointcloud $P_t$ and each point $p_s$ in the semantic pointcloud, we find the nearest neighbors using the dual k-d tree structure. This can be expressed as: 
\begin{equation}
{
(p_t, p_s) = \underset{q_t \in P_t, q_s \in P_s}{\operatorname{arg\,min}} \|q_t - q_s\|}
\end{equation}

where $\|q_t = q_s\|$ denotes the Euclidean distance. The dual k-d tree structure allows this search to be performed with average time complexity of $O(\log m + \log n)$ \cite{skrodzki2019kd} where $m$ and $n$ are the numbers of points in the geometric and semantic point clouds respectively. This efficiency is essential for the navigation requirement of processing large-point cloud data in real-time. The result of this step is a set of point pairs $(p_t, p_s)$ where points from both point clouds are associated based on their proximity in the common map frame.

Once the semantic and geometric data are associated, we calculate a traversability cost for each point pair using a cost function that combines both types of information: 
\begin{equation}
C(p)=C_s(p)w + C_g(p)(1-w) 
\end{equation}
where $C(p)$ is the total cost for point $p$, $C_s(p)$ is the semantic cost based on the class label of point $p$, $C_g(p)$ is the geometric cost derived from terrain analysis and $w$ is a weighting parameter. The total cost $C(p)$ is then associated with the $(x,y,z)$ coordinates of $p_s$ to form a pointcloud containing traversability cost. 

\begin{figure}[b]
\centering
\includegraphics[width=0.75\linewidth]{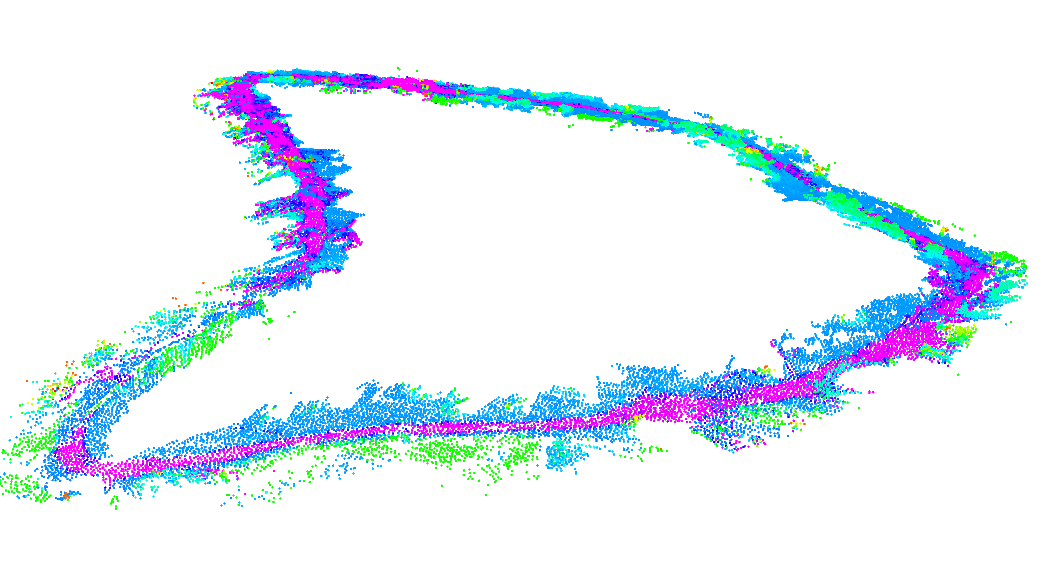}
\caption{Traversability map of Guthrie Loop at the West Virginia University Core Arboretum. Built using weight value of 0.75.}
\label{fig:traversability_map}
\end{figure}

To maintain spatial coherence in the global map, a two-step registration process is employed for each new fused pointcloud. First, odometry data from a mapping algorithm is used to compute an initial guess between the current and previous point clouds. Initial alignment is then refined using the iterative closest point (ICP) algorithm \cite{924423}. Given two point sets $P = {p_1, ..., p_n}$ and $Q = {q_1, ..., q_m}$, ICP iteratively minimizes the following error function: 
\begin{equation}
E(R, t) = \frac{1}{N_p} \sum_{i=1}^{N_p} |p_i - (Rq_i + t)|^2
\end{equation}
where $R$ is the rotation matrix, $t$ is the translation vector, and $N_p$ is the number of corresponding point pairs. In our application, $P$ represents the newest pointcloud, and $Q$ represents the global pointcloud. This iterative refinement helps to correct any inaccuracies in the initial odometry-based estimate and ensure a more precise alignment of the new pointcloud with the global map. The global traversability map is constructed incrementally by integrating newly registered pointclouds. The resulting traversability map, such as the one in Figure \ref{fig:traversability_map}, provides an environment representation that is the foundation for our trail navigation. 




\subsection{Waypoint Selection and Path Planning}

A hierarchical approach is used for navigation. A traversability-based global planner using the Rapidly-exploring Random Tree Star (RRT*)\cite{karaman2011incremental}  algorithm generates intermediate waypoint goals for the robot. Each point in the traversability map has a weight associated that ranges from $0.0$ being non-traversable to $1.0$ being the most traversable. The main difference from traditional RRT* implementation is that this algorithm uses the probability distribution from the traversability map as the sampling method for selecting candidate waypoints, instead of using a random sample formulation, biasing the tree exploration towards more traversable points. This algorithm also considers both the traversability of the terrain and the distance to generate the cost associated to the tree growth and rewiring steps

The configuration space $\mathcal{C} \subset \mathbb{R}^3$ is defined with initial state $(x_{\text{init}}, y_{\text{init}},z_{\text{init}})$ and goal state $(x_{\text{goal}}, y_{\text{goal}},z_{\text{goal}})$. The algorithm iteratively adds new nodes. Points $x_{\text{candidate}} \in {TraversabilityMap}$ are sampled using the probability associated with the traversability distribution. The nearest vertex $x_{\text{nearest}} \in V$ to a candidate sample $x_{\text{candidate}}$ is found using the Euclidean distance metric. Then, if within a boundary, the candidate node  $x_{\text{candidate}}$ is added to the tree.

The cost of traversing from a node $x$ to its parent $x_{\text{parent}}$ is computed as:
\begin{multline}
c(x) = c(x_{\text{parent}}) + d(x_{\text{parent}}, x) \\ + \begin{cases} 
\frac{W}{\frac{T(x_{\text{parent}}) + T(x)}{2}} & \text{if } T(x_{\text{parent}}) \neq 1 \text{ or } T(x) \neq 1 \\
0 & \text{if } T(x_{\text{parent}}) = 1 \text{ and } T(x) = 1 
\end{cases}
\end{multline}

Where $W$ is the traversability weight and $T(x)$ is the traversability value of node $x$. The inclusion of the traversability weight penalizes paths that traverse through less desirable areas. The traversability cost component is only applied when either of the nodes has a traversability value of less than 1. Nodes with a traversability value that is 0 are considered to be within a collision space and thus are not added to the tree. A rewiring step occurs each time a node is added. After evaluating the path, our global planner publishes the next waypoint goal for the local planner sequentially in a receding horizon manner, based on the current position of the robot. The local planner and path-following algorithm FALCO \cite{zhang2020falco} is used by the robot for navigation. This is a lattice-based local planner based on pre-computed robot kinematics.

\section{Experiments and Results}
\label{experiments_results}

\begin{table*}
\centering
\caption{
Effects of semantic vs. geometric weighting in traversability estimation on hiking trail navigation.}
\label{tab:method_comparison}
\small
\begin{adjustbox}{width=0.8\textwidth}
\begin{tabular}{|l|*{12}{c|}}
\hline
\multirow{3}{*}{\parbox{1.5cm}{Weight value }} & \multicolumn{6}{c|}{Path 1} & \multicolumn{6}{c|}{Path 2} \\
\cline{2-13}
 & \multicolumn{2}{c|}{\parbox{2cm}{\centering Time (s)}} & \multicolumn{2}{c|}{\parbox{2cm}{\centering Distance Traveled (m)}} & \multicolumn{2}{c|}{\parbox{2cm}{\centering Percentage of Drive on Trail (\%)}} & \multicolumn{2}{c|}{\parbox{2cm}{\centering Time to Traverse (s)}} & \multicolumn{2}{c|}{\parbox{2cm}{\centering Distance Traveled (m)}} & \multicolumn{2}{c|}{\parbox{2cm}{\centering Percentage of Drive on Trail (\%)}} \\
\cline{2-13}
 & Mean & Std Dev & Mean & Std Dev & Mean & Std Dev & Mean & Std Dev & Mean & Std Dev & Mean & Std Dev \\
\hline
0 & 118.98 & 6.032 & 157.33 & 7.999 & 24.93 & 2.64 & 142.7 & 9.056 & 186.27 & 4.748 & 18.768 & 3.061 \\
\hline
0.25 & 119.14 & 6.764 & 159.94 & 8.209 & 32.161 & 1.275 & 139.02 & 12.686 & 178.87 & 29.495 & 19.606 & 2.621 \\
\hline
0.5 & 114.14 & 3.112 & 140.8 & 1.695 & 45.69 & 4.414 & 108.47 & 6.069 & 146.2 & 5.427 & 31.83 & 3.691 \\
\hline
0.75 & 128.23 & 7.491 & 146.35 & 4.554 & 52.57 & 3.947 & 119.89 & 10.495 & 144.46 & 8.714 & 46.57 & 5.903 \\
\hline
1 & 123.25 & 7.701 & 151.28 & 4.455 & 67.47 & 3.531 & 193.64 & 23.27 & 225.1 & 28.18 & 64.65 & 3.035 \\
\hline
\end{tabular}
\end{adjustbox}
\end{table*}

Experiments were performed to validate our approach and analyze how the fusion of semantic and geometric data influences traversability estimation and trail navigation. Our primary goal was to understand the trade-offs between trail adherence, traversal time, and distance traveled as we adjusted the weighting of semantic and geometric information in our fusion method. First, we conducted experiments in a simulation environment, where our method was tested on various virtual trail scenarios. This environment allowed us to systematically vary the weighting between semantic and geometric information, providing insights into the method's behavior under controlled conditions. Following the experiments in the simulation we conducted the real-world test using a Clearpath Robotics Husky A200 robot on local hiking trails. These field tests were important in verifying the applicability and performance of our approach in natural outdoor environments, where factors such as varying lighting conditions and complex terrain come into play.

\subsection{Experiments in simulation environment}

\begin{figure}
\centering
\includegraphics[width=0.75\linewidth]{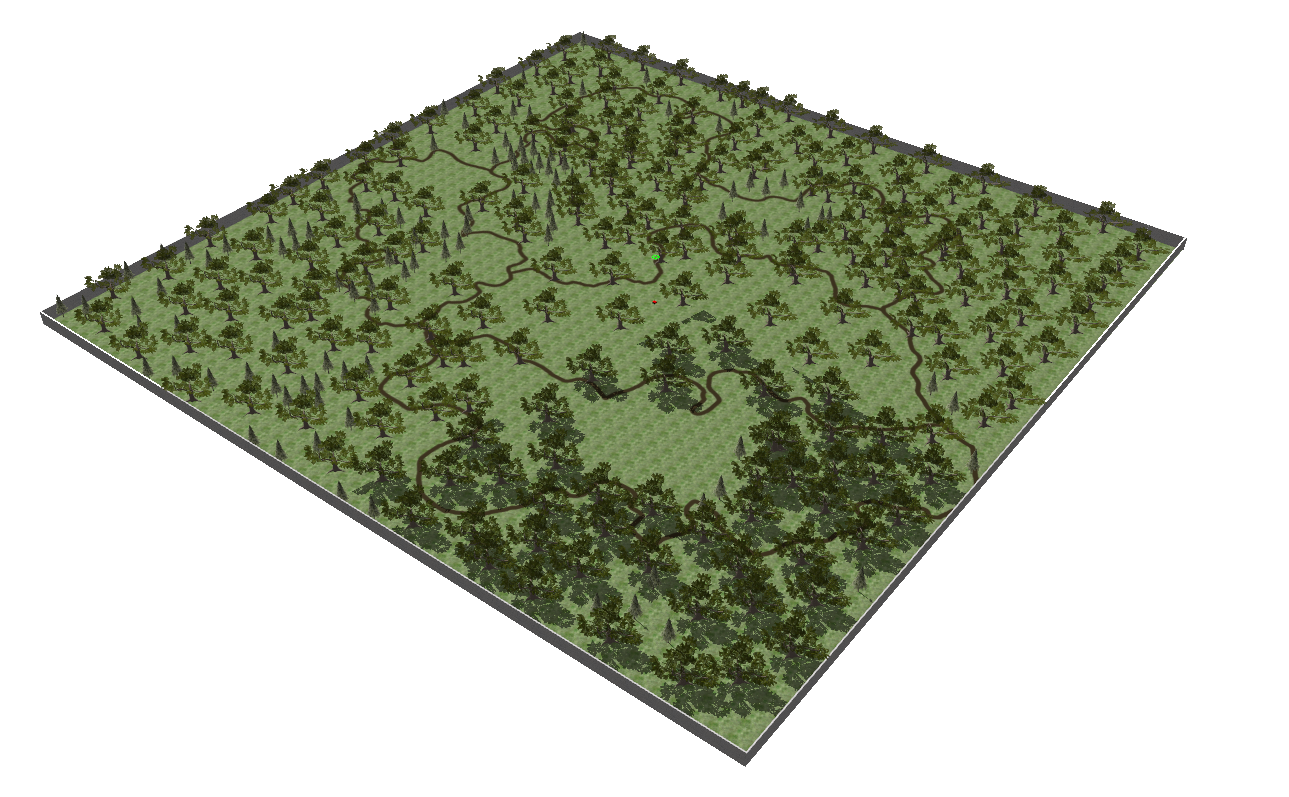}
\caption{Hiking-trail network in simulation.} 
\label{fig:simulation}
\end{figure}

A simulation environment was developed in Gazebo software to evaluate the presented modules. This simulation can also be used as a benchmark for future methods for hiking trail navigation. The simulation is built upon the CMU exploration environment simulation\cite{cao2022autonomous}. Hiking trails and various obstacles were added to their forest environment to replicate hiking conditions, as shown in Figure \ref{fig:simulation}.

We conducted experiments within this simulation to investigate the impact of different weightings between semantic and geometric information. We focused on three key metrics: distance traveled, time to traverse, and percentage of drive on trail terrain. In the context of hiking trail navigation, we consider trail adherence to be a crucial metric. Primarily, it minimizes impact on surrounding ecosystems by preventing soil erosion, vegetation damage, and wildlife habitat disturbance that can result from off-trail travel. Designated trails also offer more predictable terrain, reducing the risk of encountering unexpected hazards. Furthermore, trails often consist of packed dirt, rock, or clay, providing a path of least resistance compared to traversing through grass or dense vegetation. This can be particularly beneficial for robotic platforms over extended periods of operation.

Two distinct paths were chosen within our simulated hiking trail network, each presenting different challenges such as sharp turns, rocks and trees in the path. For each path, we performed 5 runs with 5 different weights starting at 0 and incrementing by 0.25 up to 1, where 0 means traversability is solely based on geometric information and is our benchmark comparison, and 1 means traversability is solely based on semantic information. This experimental design allow us to observe how changing the balance between geometric and semantic information affected the waypoint selection module and the overall trajectory of the robot in trail scenarios. 

Table \ref{tab:method_comparison} summarizes the quantitative results of our trials. Weights under 0.5 resulted in worse performance in all three metrics, with geometric-only traversability estimation providing the worst result. The high traversal distances and times and low percentages of time on trail indicate that these weights are inadequate for autonomous hiking trail navigation applications. In the 0.5-1 range we observe that with semantic weighting increasing in this range, the percentage of drive on the trail is improved, but at the cost of longer overall traversal times and distance, indicating a clear trade-off between trail adherence and traversal distance and time. 

\subsection{Hiking trail}
To demonstrate the effectiveness of our method we deployed BrambleBee \cite{ohi2018design}, a Clearpath Robotics Husky A200 robot, on the Arboretum trail at West Virginia University, as shown in \ref{fig:robot_on_trail}. BrambleBee was equipped with a Velodyne HDL-32E LiDAR and ZED 2i stereo camera, mirroring our simulation setup. We utilized LIO-SAM \cite{shan2020lio} algorithm as our localization solution for all demonstrations.

\begin{figure}
\centering
\includegraphics[width=1.0\linewidth]{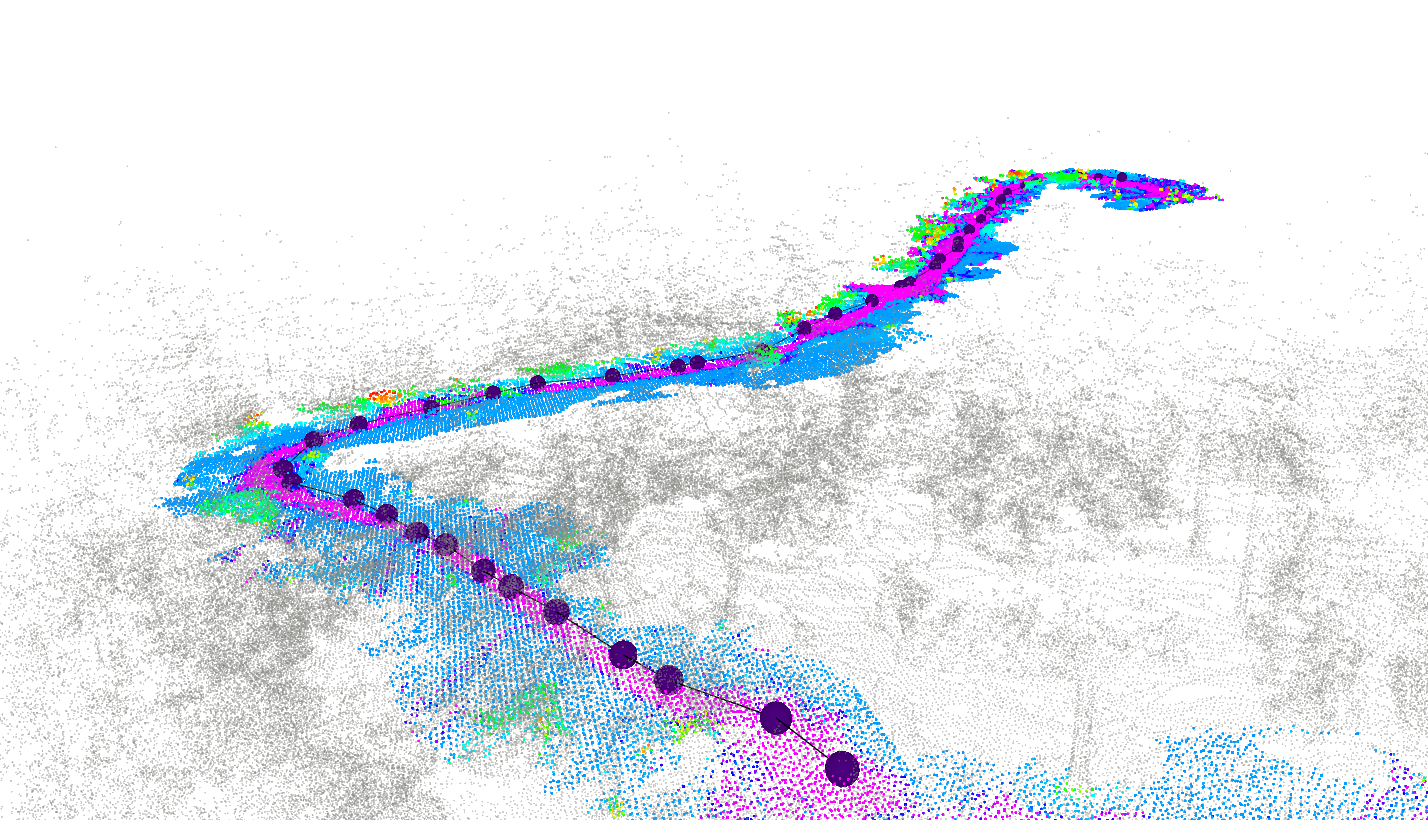}
\caption{Traversability map from WVU Core Arboretum Hiking Trail, including generated waypoints for navigation.} 
\label{fig:full_trail}
\end{figure}

For our demonstration, we chose a section of the trail with a significant change in elevation and many geometric hazards that would not be traversable for the Husky if traversing only on the trail. This path was chosen not only to challenge our system's ability to effectively identify and avoid on-trail obstacles but also to test the system’s decision-making process in finding alternative routes. We first teleoperated the robot over the chosen path, building the traversability map with a weight value of $w=0.75$ as the robot traversed. This traversability map was then passed to the waypoint selection module. The generated intermediate waypoints along the path and the traversability map are shown in Figure \ref{fig:full_trail}. Our system successfully identified impassable objects and routed the robot through a grass-covered area to avoid them, only navigating back to the trail once a clear path became available. While the detour resulted in a longer travel distance, it significantly enhanced safety and demonstrated the system’s ability to make intelligent decisions in complex scenarios.

The simulations and the field experiments are demonstrated in this video: \href{https://youtu.be/pKL58-a_eTI}{$https://youtu.be/pKL58-a_eTI$}, and validated our method's effectiveness and provided valuable insights into the system’s performance in real-world conditions. The successful navigation of both a winding trail and an obstructed path highlights the versatility and reliability of our autonomous hiking trail navigation system.



\section{Conclusion and Future Work}
\label{conclusion}

This work presented an approach to autonomous hiking trail navigation, addressing the unique challenges posed by unstructured natural environments. Our system includes traversability estimation and intelligent waypoint selection, leveraging both geometric and semantic information to enhance navigation in outdoor settings. This allows for an autonomous vehicle to adhere to the trail under normal conditions while taking an off-trail detour if an efficient shortcut is identified or if the vehicle encounters on-trail hazards.

The development and testing of our system, both in simulated environments and on real hiking trails, have provided valuable insights into the challenges of autonomous hiking trail navigation. Our integrated approach demonstrated the potential for autonomous systems to navigate complex, unstructured environments more effectively than traditional methods relying solely on geometric or semantic information.

Future work includes expanding the semantic classification dataset to improve segmentation accuracy, particularly for underrepresented classes. To enhance generalization, data from different seasons of the year and additional hiking trails with varying scenery could be added to the existing dataset. Further optimization of the waypoint selection algorithm can improve the balance between trail adherence and taking off-trail detours.

\bibliographystyle{IEEEtran}
\bibliography{bibliography}

\end{document}